\documentclass[a4paper, 10pt, conference]{ieeeconf}      
\PassOptionsToPackage{table,xcdraw}{xcolor}
\let\labelindent\relax
\usepackage{enumitem}
\usepackage{FG2025}
\usepackage{amsmath,booktabs}
\usepackage{amssymb,cite}
\usepackage{diagbox}
\usepackage{multirow}
\usepackage{courier}
\usepackage{graphicx}
\usepackage[table]{xcolor}
\usepackage{makecell}
\usepackage{algorithm}
\usepackage{algpseudocode}
\usepackage{enumitem}
\usepackage[pagebackref,breaklinks,colorlinks]{hyperref}
\newcommand{\hl}[1]{\textcolor[rgb]{0,0,0}{#1}}

\newcommand{\hlg}[1]{\cellcolor[rgb]{0.85,0.85,0.85}#1}

\FGfinalcopy 

\IEEEoverridecommandlockouts                              
\def\FGPaperID{88} 

\title{\LARGE \bf
SelfMAD: Enhancing Generalization and Robustness in Morphing Attack Detection via Self-Supervised Learning}

\author{\parbox{16cm}{\centering
    {\large Marija Ivanovska$^1$, Leon Todorov$^2$, Naser Damer$^3$, Deepak Kumar Jain $^4$, Peter Peer$^2$, Vitomir Štruc$^1$}\\
    {\normalsize
    $^1$ Faculty of Electrical Engineering, University in Ljubljana, Slovenia\\
    $^2$ Faculty of Computer and Information Science, University in Ljubljana, Slovenia\\
    $^3$  Dalian University of Technology, China\\
    $^4$  Fraunhofer Institute for Computer Graphics Research, Germany}\\}
    \thanks{The research presented in this work was supported in parts by the ARRS Research Programme P2-0250(B) "Metrology and Biometric Systems", and the ARRS Research Project J2-50065 "DeepFake DAD".}
}
\renewcommand{\baselinestretch}{0.99}

\begin{document}

\ifFGfinal
\thispagestyle{empty}
\pagestyle{empty}
\else
\author{Anonymous FG2025 submission\\ Paper ID \FGPaperID \\}
\pagestyle{plain}
\fi
\maketitle

\begin{abstract}

With the continuous advancement of generative models, face morphing attacks have become a significant challenge for existing face verification systems due to their potential use in identity fraud and other malicious activities. Contemporary Morphing Attack Detection (MAD) approaches frequently rely on supervised, discriminative models trained on examples of bona fide and morphed images. These models typically perform well with morphs generated with techniques seen during training, but often lead to sub-optimal performance when subjected to novel unseen morphing techniques. While unsupervised models have been shown to perform better in terms of generalizability, they typically result in higher error rates, as they struggle to effectively capture features of subtle artifacts. To address these shortcomings, we present SelfMAD, a novel self-supervised approach that simulates general morphing attack artifacts, allowing classifiers to learn generic and robust decision boundaries without overfitting to the specific artifacts induced by particular face morphing methods. Through extensive experiments on widely used datasets, we demonstrate that SelfMAD significantly outperforms current state-of-the-art MADs, \hl{reducing the detection error by more than $\mathbf{64\%}$ in terms of EER when compared to the strongest unsupervised competitor, and by more than $\mathbf{66\%}$, when compared to the best performing discriminative MAD model}, tested in cross-morph settings. The source code for SelfMAD is available at \href{https://github.com/LeonTodorov/SelfMAD }{https://github.com/LeonTodorov/SelfMAD}.

\end{abstract}

\vspace{-5pt}
\section{Introduction}
Automatic face recognition systems (FRSs)~\cite{grm2018strengths} are commonly employed to verify an individual's identity by matching their face image with the corresponding data stored in the system's database. Although these systems are widely used and generally very accurate, they are vulnerable to certain types of attacks representing manipulated data. A notable example are \textit{face morphing attacks}~\cite{Bush_FRS_under_MA_ieee_access_2019, Scherhag2017,Ferrara2016}, created by blending the facial features of two or more individuals. The resulting morphed image can then be used to falsely authenticate any person whose facial attributes were utilized in the morphing process.

In the age of big data and major advancements in generative models, the widespread availability of open-source morphing techniques have made it nearly effortless to create realistic, high-quality morphed face images~\cite{damer_MorDIFF_IWBF_2023, Zhang_diffMorpher_2024_CVPR}. The automatic detection of face morphing attacks is therefore critical for preventing illegal activities~\cite{nalini_book_manipulaed_faces_elsevier_2023} such as identity theft, personal document frauds, social engineering attacks etc. In recent years, the threat posed by morphed image attacks has been predominantly addressed with the development of powerful \textit{morphing attack detection (MAD)} methods. In most cases, they rely on \textit{supervised, contrastive learning}, optimizing models to differentiate between bona fide images and examples of known morphing attacks~\cite{caldeira_IDistill_EUSIPCO_2023,Boutros_MixFaceNet_2021, Ramachandra_Inception_2020}. Although such an approach tends to be highly accurate when tested on attacks encountered during  training, these methods \textit{(i)} fail to detect morphs generated by unfamiliar or novel attacking techniques~\cite{damer_MorDIFF_IWBF_2023}; and \textit{(ii)} their performance usually declines, when applied to data from unfamiliar sources, due to domain shifts~\cite{Damer2022_OC_MAD_SPA}.

\begin{figure}[!tb]
    \begin{center}
        \includegraphics[width=0.93\linewidth]{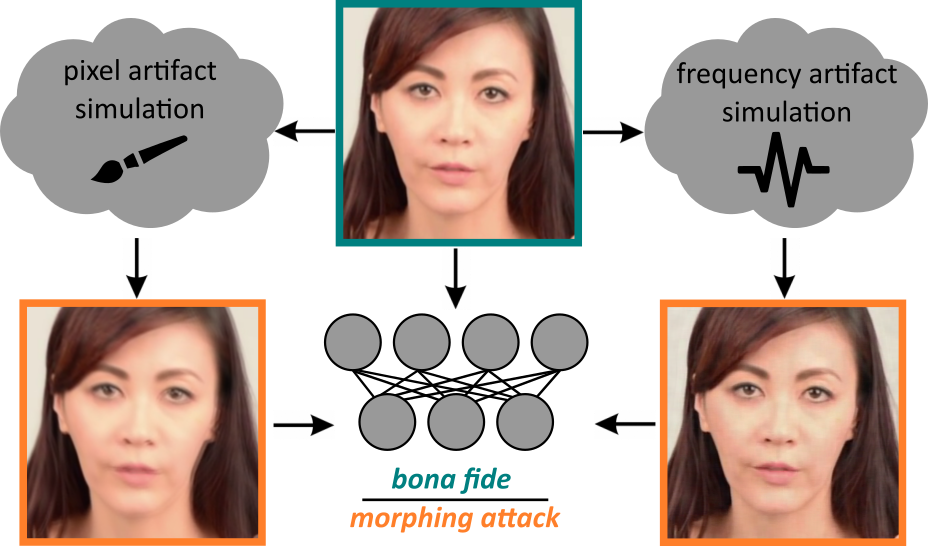}\vspace{-1mm}
        \caption{We propose \textbf{SelfMAD}, a \textbf{Self}-supervised \textbf{M}orphing \textbf{A}ttack \textbf{D}etection method, that learns to \textbf{detect morphed faces by replicating common artifacts} of various widely used morphing techniques. Pixel space manipulations simulate artifacts typical for image-level morphing techniques, while frequency space manipulations reproduce the fingerprints of latent-space morphing techniques. Self-MAD is robust and generalizes effectively, without overfitting to specific face morphing attack examples.} \label{fig:teaser}
    \end{center}
\vspace{-6mm} \end{figure}

To overcome the limitation in the generalization capabilities of supervised MADs, some researchers have investigated the use of \textit{unsupervised, one-class models} \cite{Damer2019_OC_MAD,Ibsen2021_OC_MAD}. Instead of relaying on contrastive learning, their objective is to capture only characteristics of bona fide training images and detect morphing attacks as out-of-distribution samples. While one-class models are expected to generalize better to previously unseen morphs, they frequently: \textit{(i)} mistakenly classify legitimate samples as attacks, particularly when the bona fide training data does not fully capture the diversity of real-world scenarios~\cite{Damer2022_OC_MAD_SPA}; \textit{(ii)} struggle with high sensitivity to the quality and quantity of the bona fide images used during training, reducing the models' robustness to varying conditions~\cite{ivanovska2023mad_ddpm}; and \textit{(iii)} fail to learn distinctive bona fide features that are absent in manipulated data, which hinders the MAD models' ability to identify morphing 
attacks as samples that do not fit the modeled patterns~\cite{Damer2019_OC_MAD}.

Recently, \textit{self-supervised learning} has emerged as a promising alternative to the fully unsupervised training of one-class models. Unlike unsupervised methods, self-supervised learning typically utilizes synthetically generated samples or perturbations, to enhance the model's ability to detect subtle or complex anomalies in various forms of manipulated data. In the context of image manipulation detection, self-supervised models have been successfully applied to tasks such as anomaly detection in industrial images~\cite{Zavrtanik_DRAEM_2021_ICCV,Li_CutPaste_2021_CVPR, KIM_ssl_ad_stylegan_2023}, detection of adversarial attacks~\cite{li_ssl_adversarial_attack_eccv_2024, naseer_ssl_adversarial_robust_cvpr_2020,jain_faceguard_fg_2023},  presentation attack detection~\cite{Patel_ssl_pad_IJCB_2020, Chugh_ssl_fingerprint_spoof_tifs_2021,kim_ssl_fingerprint_pad_mipr_2019}, deepfake detection~\cite{li_DSP_FWA_cvprw_2019,Shiohara_SBI_CVPR_2022,Li_FaceXray_cvpr_2020}, detection of AI-synthesized images~\cite{wang_ssl_freq_aug_icme_2023, coccomini_freq_injection_2024, lu_ai_synthesized_faces_2024} etc. Despite these advancements, the full potential of self-supervised learning methods applied to the task of morphing attack detection has yet to be explored.

In this paper, we address the challenges of both supervised and unsupervised approaches by framing morphing attack detection as a self-supervised task. \hl{As a result, we make the following main contributions in this work:} 
\begin{itemize}[noitemsep,leftmargin=*]
\item \hl{We propose \textbf{SelfMAD}, a novel \textit{\textbf{Self}-supervised \textbf{M}orphing \textbf{A}ttack \textbf{D}etection} method, designed to detect attacks by simulating common face morphing artifacts present across various types of manipulations (Fig.~\ref{fig:teaser}).}
\item \hl{Through extensive experiments, we demonstrate that our approach enhances the detection generalization to previously unseen morphing attacks, reducing the risk of overfitting to specific morphing techniques.} 
\item \hl{We perform a rigorous comparison study against widely used and highly competitive MAD models, showing that SelfMAD offers a robust solution, effectively balancing the strengths of both supervised and unsupervised methods.} 
\item \hl{We conduct an in-depth ablation study to assess the impact of SelfMAD components.}
\end{itemize}

\vspace{-5pt}
 \section{Related work}\label{sec:related_work}

In this section, we first discuss different widely used techniques for generation of face morphing attacks. Next, we present a brief overview of the studies on morphing attack detection (MAD). Finally, we review self-supervised learning (SSL) methods applied for detection of out-of-distribution samples, providing a background of our research work. For a more in-depth coverage of these topics, readers are referred to some of the excellent surveys
~\cite{Raghavendra_MAD_survey_2021,Bush_FRS_under_MA_ieee_access_2019,Hojjati_SSL_AD_2024}.

\vspace{1mm}\noindent\textbf{Morphing Attack Generation.} Face morphing attack generation techniques are generally categorized into traditional and deep learning-based methods. Traditional approaches typically operate in the pixel space and involve three main steps: aligning corresponding face features from two images, warping them to match geometrically, and blending the warped images to merge color values~\cite{scherhag2019face}. The alignment relies on detected facial landmarks, and different methods may use various warping techniques~\cite{Schaefer_least_squares_2006,Hildebrandt_morphing_iwbf_2017,Makrushin_morphing_VISIGRAPP_2017, Scherhag_morphing_iwbf_2017}. However, this process can introduce misaligned pixels, leading to artifacts and ghost-like images that are often noticeable. To address these issues, post-processing steps such as image smoothing, sharpening, edge correction, and histogram equalization are commonly applied to enhance image quality and reduce artifacts~\cite{Seibold_morphing_IWDW_2017, Weng_interpolation_forum_2013}.

With the advent of advanced deep learning-based generative models, recent approaches have significantly improved the quality of morphed face images compared to traditional landmark-based methods. These modern techniques typically involve embedding two face images into the latent space of a generative model, performing vector interpolation to create a morphed image, and then decoding this vector back into pixel space. While Generative Adversarial Networks (GANs) are commonly used for this purpose~\cite{Damer_MorGAN_BTAS_2018,Venkatesh_StyleGAN_morphs_IWBF_2020,Zhang_MIPGAN_2021}, newer methods also incorporate diffusion-based networks to achieve even greater image quality and realism~\cite{damer_MorDIFF_IWBF_2023,Blasingame_improved_morphing_diffusion_tbiom_2024, Blasingame_greedy_IJCB_2024}. Despite their high level of realism and difficulty to detect, morphs produced by these generative models still exhibit detectable irregularities in texture or frequency patterns, which can reveal their manipulated nature~\cite{Corvi_synthetic_image_properties_CVPRW_2023}.

\begin{figure*}[!th]
    \begin{center}
        \includegraphics[width=0.98\linewidth]{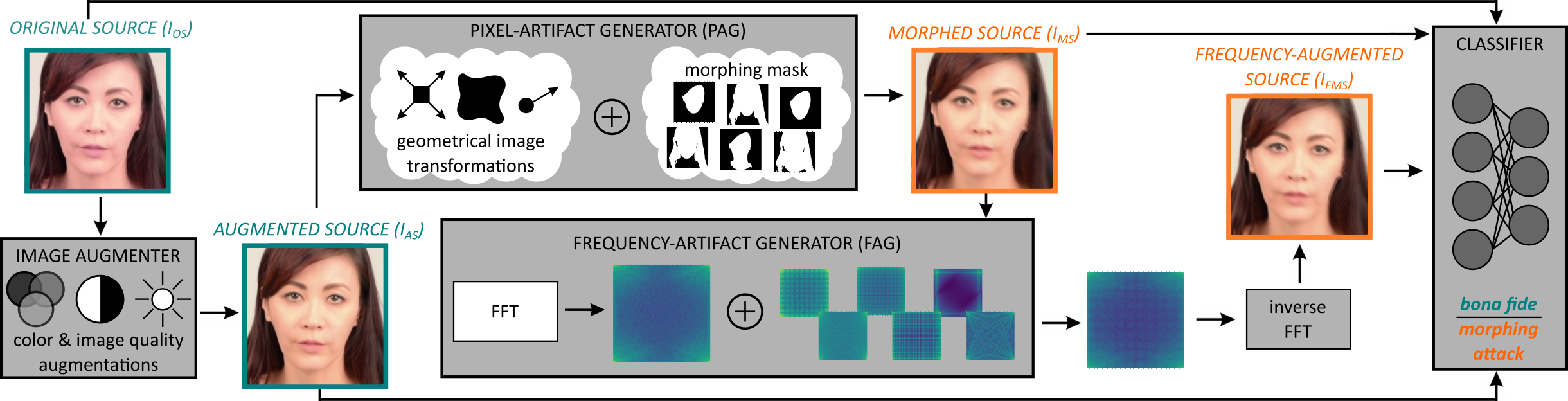}
        \caption{\textbf{Overview of Self-MAD}, a self-supervised morphing attack detection method that learns to detect morphed faces using a set of bona fide images and simulated morphing attacks. The model consists of four key components: \textit{i) pixel augmenter}, to simulate subtle visual variations in real-world images; \textit{ii) pixel artifact generator} that mimics artifacts typical for image-level morphing methods; \textit{iii) frequency artifact generator} which reproduces frequency fingerprints associated with advanced, latent-level morphing techniques; and \textit{iv) classifier} differentiate between genuine and manipulated samples.}\label{fig:model}
    \end{center}
\vspace{-6mm} \end{figure*}

\vspace{1mm}\noindent\textbf{Morphing Attack Detection.} Existing morphing attack detection (MAD) models are in general categorized into single-image (S-MAD) and differential (D-MAD) approaches. S-MAD models analyze face morphs individually, without comparing them to other images, whereas D-MAD models compare manipulated samples to a reference image. D-MADs are typically highly accurate in closed-set scenarios, while S-MADs are primarily employed to detect attacks when there is no prior knowledge of the subjects' identities. In this section, we focus our literature review exclusively on S-MADs, as they are most closely related to our work.

Regardless of the face morphing technique employed, the resulting morphs typically exhibit image irregularities such as noise, pixel discontinuities, distortions, spectrum discrepancies, and other visual artifacts. Early MAD approaches aimed to detect these irregularities using hand-crafted techniques, including photo-response non-uniformity (PRNU) noise analysis~\cite{Scherhag_PRNU_2019}, reflection analysis~\cite{Seibold_reflection_2018}, or texture-based descriptors like LBP~\cite{Ojala_1996_LBP}, LPQ~\cite{Ojansivu_LPQ_2008}, or SURF~\cite{Makrushin_SURF_2019}. While these methods produced promising results, their ability to generalize across different scenarios was limited~\cite{Damer2019_OC_MAD}.

More sophisticated MADs take advantage of the capabilities of data--driven, deep learning algorithms~\cite{SYN_MAD_2022}. Raghavendra \textit{et al.}~\cite{Raghavendra_transfer_2017}, for instance, were amongst the first proposing pretrained deep models as a supervised approach to the detection of morphing attacks. In their work, morphs were detected with a simple, fully-connected binary classifier, fed with fused VGG19 and AlexNet features, pretrained on ImageNet. Similarly, Wandzik \textit{et al.}~\cite{Wandzik_FRS_2018}, achieved high detection performance with features extracted from a general-purpose face recognition systems (FRSs), fed to an SVM, while another study of Ramachandra \textit{et al.}~\cite{Ramachandra_Inception_2020} utilized Inception models for the same purpose. Damer \textit{et al.}~\cite{Naser_PW_MAD_2021} on the other hand argued that pixel-wise supervision, where each pixel is classified as a bona fide or a morphing attack, is superior, when used in addition to the binary, image-level objective. MixFaceNet~\cite{Boutros_MixFaceNet_2021} by Boutros \textit{et al.}, another highly efficient deep learning architecture inspired by mixed depthwise convolutions, demonstrates even better results~\cite{Damer_SMDD_2022}, by capturing different levels of attack cues through differently-sized convolutional kernels. A more complex approach is proposed by Neto \textit{et al.}, who incorporate a regularization term during the training of their model OrthoMAD, whose goal is to disentangle identity features for more robust morphing attack detection. The follow-up method, IDistill further improves MAD results by introducing more efficient feature disentanglement and adding interpretability. All these supervised approaches however typically experiance a significant decline in their performance, when applied to images that do not fit the training distribution.  

In some more recent studies, researchers have tried to improve the generalization capabilities of the MAD techniques, by proposing unsupervised, one-class learning models trained exclusively on bona fide samples. Different from the supervised techniques discussed above, Damer \textit{et al.}~\cite{Damer2019_OC_MAD}, for example, were among the first to achieve significant performance generalization on unseen attacks with two different one--class methods, i.e. a one-class support vector machine (OCSVM) and an isolation forest (ISF). Comparable generalization capabilities were later demonstrated in~\cite{Ibsen2021_OC_MAD}, where Ibsen \textit{et al.} explored the use of a Gaussian Mixture Model (GMM), a Variational Autoencoder (VAE) and Single-Objective Generative Adversarial Active Learning (SO-GAAL) in addition to an OCSVM. A more advanced approach was proposed by, Fang \textit{et al.}~\cite{Damer2022_OC_MAD_SPA} who enhance an unsupervised convolutional autoencoder with self-paced learning (SPL). With this approach, the model neglects suspicious unlabeled training data, widening the reconstruction error gap between bona fide samples and morphing attacks. Ivanovska \textit{et al.}~\cite{ivanovska2023mad_ddpm} achieve further improvement in detection performance by deploying diffusion models to learn the distribution of bona-fide images. Different from reconstruction techniques, Fu \textit{et al.}~\cite{damer_iqa_IETB_2022} measure the authenticity of images through image quality estimation. Nevertheless, learning distinctive image features in a one-class setting is still a challenging task, frequently leading to misclassification of realistic morphs with subtle inconsistencies.  

\vspace{1mm}\noindent\textbf{Self-Supervised Learning for Anomaly Detection.} Self-supervised learning has recently emerged, as a promising alternative that tackles the challenges of out-of-distribution sample detection, by combining the strengths of both supervised and unsupervised approaches. Unlike fully supervised methods, which depend heavily on labeled data, and unsupervised one-class models that in contrast focus solely on bona fide samples, self-supervised techniques utilize automatically generated labels derived from the data itself. For instance, in industrial anomaly detection, self-supervised models simulate typical defects by synthesizing data representing anomalous samples~\cite{Zavrtanik_DRAEM_2021_ICCV,Li_CutPaste_2021_CVPR, KIM_ssl_ad_stylegan_2023}. In adversarial attack detection, Deb~\textit{et al.} adopt a similar approach by artificially creating challenging and diverse adversarial attacks in the image space~\cite{jain_faceguard_fg_2023}, while Li~\textit{et al.} and Naseer~\textit{et al.} apply data manipulation in the latent space. These strategies have been so far successfully applied to various biometric tasks~\cite{Patel_ssl_pad_IJCB_2020, Chugh_ssl_fingerprint_spoof_tifs_2021,kim_ssl_fingerprint_pad_mipr_2019}. Recently, a notable success has been achieved in tasks related to deepfake detection. Li~\textit{et al.} and Shiohara~\textit{et al.}~\cite{Shiohara_SBI_CVPR_2022,Li_FaceXray_cvpr_2020} for example, propose simulating typical image inconsistencies in deepfakes, where and original face has been either replaced of enhanced. Differently, Wang \textit{et al.} choose to simulate frequency artifacts instead of focusing on local image irregularities~\cite{wang_ssl_freq_aug_icme_2023}. More sophisticated algorithms simulating frequency-based irregularities have also been proposed by Coccomini~\textit{et al.}~\cite{coccomini_freq_injection_2024} and Lu~\textit{et al.}~\cite{lu_ai_synthesized_faces_2024} who demonstrate remarkable generalizability of this approach. 


\hl{Motivated by the success of the self-supervised paradigm, in this paper we explore its capabilities in the context of MAD (morphing attack detection). Unlike existing MAD methods, we augment normal data by simulating typical \textit{morphing irregularities} that manifest in both pixel and frequency domains. This newly generated data is then used to extract general and robust features for effective attack detection.}

\vspace{-5pt}
\section{Methodology}\label{sec:methodology}
We introduce \textbf{SelfMAD}, a self-supervised morphing attack detection method designed to recognize manipulated face images by searching for general inconsistencies that are common across various types of face morphs, and at the same time independent of specific face identities. The proposed method utilizes a proxy task, which simulates morphing inconsistencies through a three-stage image pre-processing pipeline that consists of: \textit{(i)} image augmentation, \textit{(ii)} pixel-artifact generation, and \textit{(iii)} frequency-artifact generation steps. The manipulated images generated by the pipeline, along with the original, unprocessed images, are then used for the training of a neural discriminator, which learns to distinguish between bona fide and altered samples, without using actual face morphs. \hl{Note that for simulating morphing inconsistencies, we use various transformations and frequency masks that have been shown to be suitable for modeling artifacts and subtle cues induced into facial images by various (GAN, diffusion, or pixel-level) image-manipulation techniques.\cite{Shiohara_SBI_CVPR_2022, larue2023seeable,corvi2023intriguing,yang2023fingerprints}}
A high-level overview of SelfMAD is shown in Fig.~\ref{fig:model}.

\vspace{1mm}\noindent\textbf{Image Augmenter.} The first component in our pipeline focuses on augmenting input images by applying a series of transformations that alter the visual appearance of the data, without changing it's underlying structure. The primary goal of this processing stage is to \textit{simulate subtle (global) visual variations} that may occur in real-world face images. Specifically, given an input bona fide image $I_{OS}$ (original source), the Image Augmenter applies a set of transformations $\psi$, to generate an augmented image $I_{AS}$ (augmented source):
\begin{equation}
  I_{AS}=\psi(I_{OS}),\label{eq:image_augmentor} 
\end{equation}
where $\psi\in\{${\texttt{RGBShift}, \texttt{HueSaturationValue}, \texttt{RandomBrightnessContrast}, \texttt{RandomDownScale}, \texttt{Sharpen}}$\}$ comprises five basic (global) image transformations. 
Color-related variations in real-world image are simulated by applying the first two functions in $\psi$, which introduce slight adjustments to the RGB and HSV values of the input sample $I_{OS}$. The third function additionally alters the brightness and the contrast of the image. To mimic real-world variations in image quality, the augmenter also applies compression at varying intensities, using either downsampling or sharpen operation that are implemented with the last two functions in $\psi$. The parameters for all five transformation functions are randomly selected within predefined ranges. After applying $\psi$, the content of the input image $I_{OS}$ remains unchanged, ensuring that the transformed output image $I_{AS}$ is still considered bona fide.

\vspace{1mm}\noindent\textbf{Pixel-Artifact Generator.} The second image processing component of SelfMAD \textit{introduces pixel space artifacts} to simulate image irregularities created by traditional, landmark-based morphing techniques (see section~\ref{sec:related_work}). To faithfully replicate the image-level morphing of two non-identical faces, the pixel artifact generator performs pixel space blending of the original source image $I_{OS}$ and the geometrically transformed version of $I_{AS}$ from the previous SelfMAD step. 
Formally, the pixel-artifact generator first creates $I_{AS}\prime$ by transforming $I_{AS}$ with set of functions $\zeta$:
\begin{equation} I_{AS}\prime=\zeta(I_{AS}),\label{eq:pixel_artifact_generator_geo_transforms} 
\end{equation}
where $\zeta\in\{${\texttt{Translation}, \texttt{ElasticTransform}, \texttt{Scaling}}$\}$ comprises three key geometrical image transformations. The first function performs only image translation in a randomly chosen direction. Therefore, it mimics the imperfect alignment of corresponding face landmarks, typical for the morphing of two different faces. The second function, the elastic transformation, is a smooth, non-linear image deformation that locally adjusts pixel positions to simulate variations in the shapes of individual face parts. This function aims to reflect the natural distinctive features of different real-world face structures. The third and last function applies image scaling, consequently adjusting the dimensions of facial features to capture potential size discrepancies. Finally, the geometrically transformed image $I_{AS}\prime$ is blended with the original source image $I_{OS}$ using a blending mask $M$:
\begin{equation}
  I_{MS} = I_{AS}\prime\odot a\cdot M + I_{OS}\odot (1-a)\cdot M,\label{eq:pixel_artifact_generator_morphing} 
\end{equation}
where $I_{MS}$ is the resulting morphed image source and $a$ is the blending factor, uniformly sampled from a set of predefined discrete values. The blending mask $M$ is a binary image, that allows the algorithm to selectively apply pixel artifacts. When all values of $M$ equal $1$, the pixel-artifact generator performs a classical pixel-wise morphing of $I_{AS}\prime$ and $I_{OS}$. Alternatively, the pixel-artifact generator can randomly choose a mask $M$ that represents a combination of two or more facial parts, segmented by a pretrained face parser. This strategy helps SelfMAD emphasize pixel irregularities in different face regions.

\vspace{1mm}\noindent\textbf{Frequency-Artifact Generator.} In the third image processing stage, the pipeline \textit{simulates frequency artifacts} commonly created by generative models that perform face morphing in their latent spaces, i.e. on face template level (see section~\ref{sec:related_work}). These morphs often appear flawless in the image space, maintaining correct semantics and visual consistency. However, they typically contain so-called frequency fingerprints that are not present in pristine images. These fingerprints can either represent structured geometrical artifacts frequently linked to generative adversarial networks (GANs) or abnormal densities in the frequency spectrum, often associated with diffusion-based models. The frequency artifact generator models these inconsistencies by first creating a random structured geometrical pattern $\Phi$, then calculating its Fast Fourier Transform $F_\Phi$:
\begin{equation}
  F_\Phi =  FFT(\Phi),\label{eq:freq_augmentor} 
\end{equation}

\noindent where the type of the pattern $\Phi$ is uniformly chosen to represent one of the following: a symmetrical grid, an asymmetrical grid, a square checkerboard, a circular checkerboard, randomly distributed squares, a set of random lines or a set of random stripes. The magnitudes $|F_\Phi|$ are then separately superimposed on the magnitudes of the Fourier transform of the image from the earlier step, i.e. the morphed source image $I_{MS}$:
\begin{equation}
    \begin{split}
        |F_{MS}|\prime= (1-k)\cdot |FFT(I_{MS})|\oplus k\cdot |F_\Phi|,
    \end{split}
\end{equation}

\noindent where $k$ is a predetermined constant of the pixel-wise summation. The resulting magnitude spectra is then transformed back to the image space, by applying the inverse FFT:
\begin{equation}
    \begin{split}
        I_{FMS}= InverseFFT(|F_{MS}|\prime, \theta(F_{MS})),
    \end{split}
\end{equation}

\noindent where $\theta$ is the phase spectra of the Fourier transform $F_{MS}$. 


\vspace{1mm}\noindent\textbf{Classifier.} The final component of SelfMAD is a classifier $\mathcal{D}$, optimized by minimizing the binary cross-entropy loss:
\begin{equation} \label{eq:bce_loss}
    \mathcal{L}_{BCE} =  -\big[y\log(\mathcal{D}(I)) + (1-y)\log(1-\mathcal{D}(I))\big],
\end{equation}

\noindent where $I$ are the input images generated at different pipeline stages ($I_{OS}, I_{AS}, I_{MS},$ and $I_{FMS}$), $y$ is the corresponding image label, which equals $0$ for bona fide images $I_{OS}$ and $I_{AS}$ and $1$ for simulated morphing attacks $I_{MS},$ and $I_{FMS}$, while $\mathcal{D}(I)$ is the predicted probability of $I$ being a bona fide or an attack. 
By leveraging a diverse set of synthetic artifacts, the classifier learns to identify general features indicative of morphing attacks rather than overfitting to characteristics of specific morphing techniques. Moreover, the usage of different blending masks during the pixel-artifact generation stage helps the classifier to focus on different regions of the image. This approach prevents the model from becoming biased toward facial parts where 
artifacts might be either more common or most obvious. 

The pseudocode of SelfMAD is given in Algorithm~\ref{alg:selfmad}.

\begin{algorithm}[htbp]
\caption{Pseudocode of SelfMAD}
\label{alg:selfmad}
\begin{algorithmic}[1]
\State \textbf{Input:} bona fide RGB images $I_{OS}^{H\times W\times3}$, constant $k$ 
\State \textbf{Output:} transformed images $I_{AS}$, $I_{MS}$, $I_{FMS}$, Classifier $\mathcal{D}$

\State \textbf{\textit{Image Augmentation}}
\For{each image $I_{OS}$}
    \State $I_{AS}\gets$ sequentially apply \{RGBShift, 
    \State HueSaturationValue, RandomBrightnessContrast,
    \State OneOf(RandomDownScale, Sharpen)\} to $I_{OS}$
\EndFor
\State \textbf{\textit{Pixel Artifact Generation}}
\For{each image $I_{AS}$}
    \State $I_{AS}\prime\gets$ sequentially apply \{Translation,
    \State ElasticTransform, Scaling\} to $I_{AS}$
    \State $M^{H\times W}\gets$ generate a binary blending mask
    \State $a\gets$ Uniform(\{0.5, 0.5, 0.5, 0.375, 0.25, 0.125\})
    \State $I_{MS}\gets$ $I_{AS}\prime\odot a\cdot M + I_{OS}\odot (1-a)\cdot M$
\EndFor

\State \textbf{\textit{Frequency Artifact Generation}}
\For{each image $I_{MS}$}
    \State $\Phi\gets$ Uniform(\{\textit{symmetrical grid}, \textit{asymmetrical grid}, 
    \State \textit{random squares}, \textit{random lines}, \textit{stripes}, 
    \State \textit{square checkerdboard}, \textit{circular checkerdboard}\})
    \State $F_\Phi\gets$ FourierTransform($\Phi$)
    \State $F_{MS}\gets$ FourierTransform($I_{MS}$)
    \State $|F_{MS}|\prime\gets (1-k)\cdot |F_{MS}|\oplus k\cdot |F_{MS}|$
    \State $I_{MAS}\gets$ InverseFourierTransform($|F_{MS}|\prime,\theta(F_{MS})$)
\EndFor

\State \textbf{\textit{Classifier training}}
\State $\mathcal{L}_{BCE} \gets \mathcal{D}(\theta,I_{OS},I_{AS},I_{MS},I_{FMS})$
\State $\omega\prime$ (updated weights of $\mathcal{D}$) $\gets$ backpropagate $\mathcal{L}_{BCE}(\omega)$
\end{algorithmic}
\end{algorithm}

\vspace{-5pt}
\section{Experimental Setup}
In this section, we first describe the datasets used in our experiments. Next, we explain the evaluation metrics and provide details about the implementation of the models.

\begin{figure}[t]
\begin{center}
\centering
  \includegraphics[width=1\linewidth]{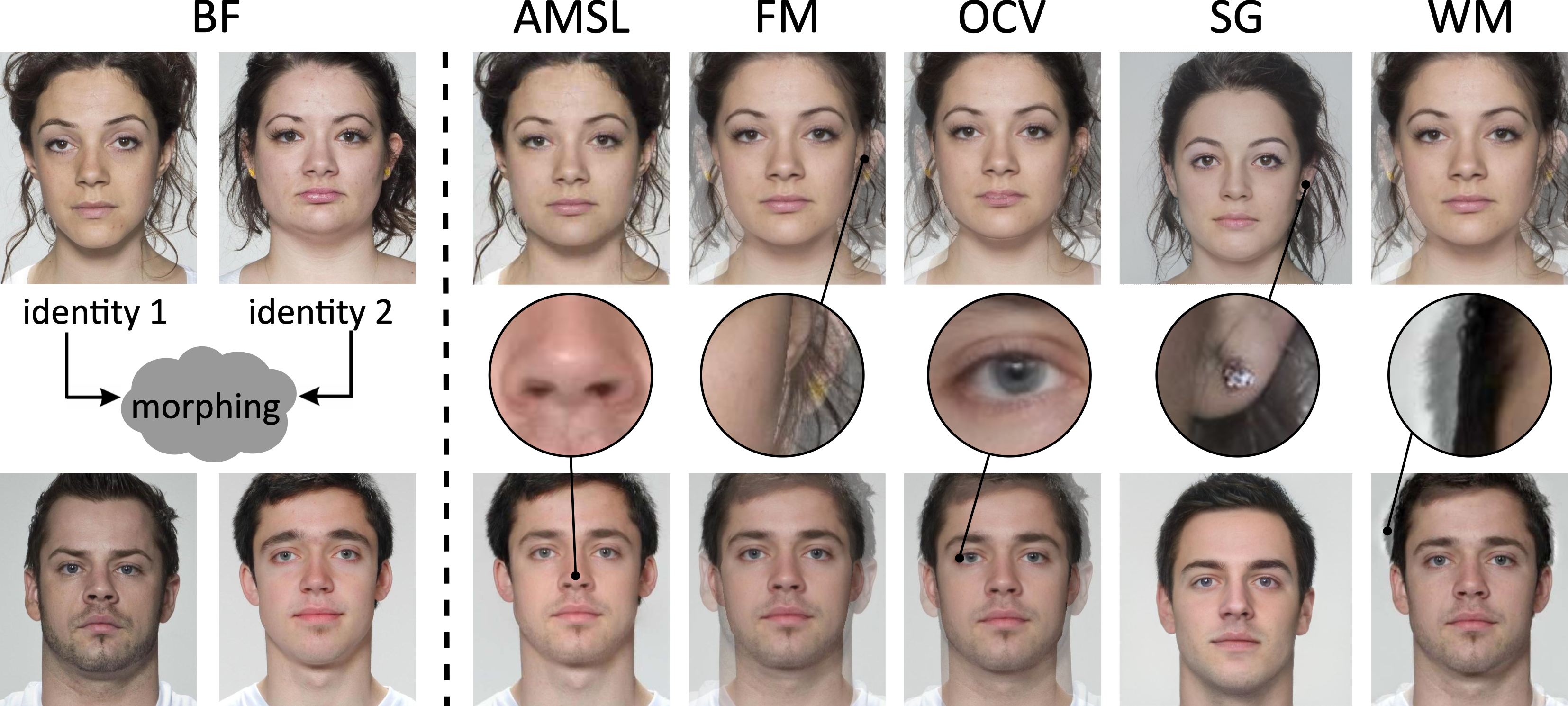}
\end{center}
\vspace{-15pt}
\caption{Selected samples from FRLL-Morphs~\cite{Sarkar2020_morphed_data}, representing bona fide (BF) images (left) and morphing attacks genertated with different morphing methods, i.e. AMSL, FaceMorpher (FM), OpenCV (OCV), StyleGAN2 (SG), and Webmorph (WM). Note typical ghosting artifacts of landmark-based morphing attacks and irregulatiries generated with StyleGAN2.\label{fig:examples}}
\vspace{-4mm} \end{figure}

\begin{table*}[!t!]
\begin{center}
\caption{\hl{\textbf{Comparison of SelfMAD against one-class MAD models}. Our SelfMAD is the top-performer across various face morphing attack types, outperforming the runner-up SPL-MAD by $10.12\%$ in terms of EER, and by $17.31\%$ and $16.1\%$ in terms of BPCER@APCER 5\% and 10\%, respectively. While SelfMAD shows slight underperformance with StyleGAN2 (SG) morphs, it significantly surpasses competitive methods in other advanced morphing attacks, i.e. Morph-PIPE, Greedy-DiM and MorCode.}}\label{tab:unsupervised}
\resizebox{\linewidth}{!}{%
\hl{
\begin{tabular}{|l c|c|c|c|c|c|c|c|c|c|c|c|c|c|c|c|c|c|c|}
\hline
\multicolumn{2}{|c|}{\multirow{3}{*}{\diagbox[width=3cm,height=0.94cm]{Test data}{Model}}} & \multicolumn{3}{c|}{\textbf{FIQA-MagFace}~\cite{damer_iqa_IETB_2022}} & \multicolumn{3}{c|}{\textbf{CNNIQA}~\cite{damer_iqa_IETB_2022}} & \multicolumn{3}{c|}{\textbf{SPL-MAD}~\cite{Damer2022_OC_MAD_SPA}} & \multicolumn{3}{c|}{\textbf{MAD-DDPM}~\cite{ivanovska2023mad_ddpm}} & \multicolumn{3}{c|}{\textbf{SBI}~\cite{Shiohara_SBI_CVPR_2022}} & \multicolumn{3}{c|}{\textbf{SelfMAD} [Ours]} \\ \cline{3-20}
& & \multirow{1}{*}{\hlg{EER}} & \multicolumn{2}{c|}{BE@AE(\%)} & \multirow{1}{*}{\hlg{EER}} & \multicolumn{2}{c|}{BE@AE(\%)} & \multirow{1}{*}{\hlg{EER}} & \multicolumn{2}{c|}{BE@AE(\%)} & \multirow{1}{*}{\hlg{EER}} & \multicolumn{2}{c|}{BE@AE(\%)} & \multirow{1}{*}{\hlg{EER}} & \multicolumn{2}{c|}{BE@AE(\%)} & \multirow{1}{*}{\hlg{EER}} & \multicolumn{2}{c|}{BE@AE(\%)} \\ \cline{4-5} \cline{7-8} \cline{10-11} \cline{13-14} \cline{16-17} \cline{19-20} 
& & \hlg{} & $5.00$ & $10.00$ & \hlg{} & $5.00$ & $10.00$ & \hlg{} & $5.00$ & $10.00$ & \hlg{} & $5.00$ & $10.00$ & \hlg{} & $5.00$ & $10.00$ & \hlg{} & $5.00$ & $10.00$ \\ 
\hline\hline 
\multirow{3}{*}{\rotatebox{0}{\textbf{FRGC-M}}} & \multicolumn{1}{|c|}{FM} & \hlg{$33.82$} & $73.79$ & $62.84$ & \hlg{$42.84$} & $75.94$ & $66.86$ & \hlg{$16.91$} & $25.39$ & $21.47$ & \hlg{$25.62$} & $95.12$ & $90.15$ & \hlg{$16.68$} & $38.07$ & $26.14$ & \hlg{$\mathbf{5.59}$} & $\mathbf{6.43}$ & $\mathbf{2.80}$\\ 
& \multicolumn{1}{|c|}{OCV} & \hlg{$33.30$} & $74.71$ & $62.52$ & \hlg{$43.15$} & $74.64$ & $66.35$ & \hlg{$20.75$} & $32.50$ & $25.42$ & \hlg{$28.22$} & $95.12$ & $90.15$ & \hlg{$15.32$} & $36.31$ & $25.10$ & \hlg{$\mathbf{2.59}$} & $\mathbf{1.14}$ & $\mathbf{0.41}$\\ 
& \multicolumn{1}{|c|}{SG} & \hlg{$14.21$} & $26.46$ & $\mathbf{17.60}$ & \hlg{$36.51$} & $70.34$ & $57.93$ & \hlg{$16.80$} & $\mathbf{26.13}$ & $21.09$ & \hlg{$\mathbf{9.02}$} & $95.12$ & $90.15$ & \hlg{$52.90$} & $97.10$ & $94.40$ & \hlg{$15.84$} & $45.23$ & $25.52$\\ \hline 
\multirow{3}{*}{\rotatebox{0}{\textbf{FERET-M}}} & \multicolumn{1}{|c|}{FM} & \hlg{$25.14$} & $61.22$ & $44.44$ & \hlg{$13.23$} & $35.17$ & $19.32$ & \hlg{$20.42$} & $40.85$ & $27.09$ & \hlg{$27.98$} & $95.27$ & $90.17$ & \hlg{$26.47$} & $60.87$ & $52.36$ & \hlg{$\mathbf{3.19}$} & $\mathbf{1.70}$ & $\mathbf{0.38}$\\ 
& \multicolumn{1}{|c|}{OCV} & \hlg{$26.14$} & $61.50$ & $43.95$ & \hlg{$20.45$} & $58.60$ & $37.23$ & \hlg{$25.71$} & $57.45$ & $45.60$ & \hlg{$31.38$} & $95.27$ & $90.17$ & \hlg{$28.73$} & $70.08$ & $60.61$ & \hlg{$\mathbf{1.13}$} & $\mathbf{0.57}$ & $\mathbf{0.38}$\\ 
& \multicolumn{1}{|c|}{SG} & \hlg{$\mathbf{12.67}$} & $\mathbf{24.63}$ & $\mathbf{15.71}$ & \hlg{$33.84$} & $79.55$ & $66.17$ & \hlg{$25.33$} & $62.06$ & $49.72$ & \hlg{$32.14$} & $95.27$ & $90.17$ & \hlg{$41.83$} & $90.55$ & $82.42$ & \hlg{$18.14$} & $46.12$ & $32.33$\\ \hline 
\multirow{5}{*}{\rotatebox{0}{\textbf{FRLL-M}}} & \multicolumn{1}{|c|}{AMSL} & \hlg{$30.94$} & $77.94$ & $66.18$ & \hlg{$21.61$} & $60.29$ & $39.22$ & \hlg{$3.26$} & $0.50$ & $0.50$ & \hlg{$27.13$} & $94.94$ & $90.02$ & \hlg{$11.76$} & $24.23$ & $16.78$ & \hlg{$\mathbf{0.99}$} & $\mathbf{0.05}$ & $\mathbf{0.05}$\\ 
& \multicolumn{1}{|c|}{FM} & \hlg{$27.99$} & $73.04$ & $57.35$ & \hlg{$19.97$} & $57.84$ & $36.76$ & \hlg{$1.03$} & $0.99$ & $0.99$ & \hlg{$10.40$} & $95.19$ & $90.38$ & \hlg{$13.73$} & $36.99$ & $26.10$ & \hlg{$\mathbf{0.00}$} & $\mathbf{0.26}$ & $\mathbf{0.17}$\\ 
& \multicolumn{1}{|c|}{OCV} & \hlg{$24.73$} & $66.18$ & $53.43$ & \hlg{$7.53$} & $11.76$ & $4.41$ & \hlg{$1.88$} & $0.50$ & $0.50$ & \hlg{$13.76$} & $95.17$ & $90.01$ & \hlg{$12.25$} & $27.85$ & $18.84$ & \hlg{$\mathbf{0.00}$} & $\mathbf{0.00}$ & $\mathbf{0.00}$\\ 
& \multicolumn{1}{|c|}{SG} & \hlg{$\mathbf{7.53}$} & $\mathbf{8.82}$ & $\mathbf{5.39}$ & \hlg{$35.92$} & $75.49$ & $68.14$ & \hlg{$14.65$} & $32.18$ & $24.75$ & \hlg{$14.32$} & $95.17$ & $90.18$ & \hlg{$44.61$} & $94.68$ & $90.92$ & \hlg{$10.34$} & $24.22$ & $12.52$\\ 
& \multicolumn{1}{|c|}{WM} & \hlg{$27.19$} & $68.14$ & $55.39$ & \hlg{$21.54$} & $46.57$ & $33.33$ & \hlg{$6.39$} & $11.39$ & $3.47$ & \hlg{$30.30$} & $95.09$ & $90.34$ & \hlg{$39.22$} & $89.93$ & $83.37$ & \hlg{$\mathbf{3.45}$} & $\mathbf{1.64}$ & $\mathbf{0.41}$\\ \hline 
\multicolumn{2}{|c|}{\textbf{Morph-PIPE}} & \hlg{$49.62$} & $91.54$ & $84.18$ & \hlg{$66.54$} & $98.83$ & $97.35$ & \hlg{$18.88$} & $33.62$ & $25.66$ & \hlg{$13.88$} & $95.14$ & $90.14$ & \hlg{$30.75$} & $92.41$ & $77.69$ & \hlg{$\mathbf{5.89}$} & $\mathbf{12.44}$ & $\mathbf{0.53}$\\ \hline 
\multicolumn{2}{|c|}{\textbf{Greedy-DiM}} & \hlg{$47.00$} & $94.61$ & $85.78$ & \hlg{$49.40$} & $96.08$ & $93.14$ & \hlg{$37.72$} & $80.69$ & $71.78$ & \hlg{$36.10$} & $95.20$ & $89.70$ & \hlg{$33.82$} & $90.60$ & $81.60$ & \hlg{$\mathbf{7.60}$} & $\mathbf{37.60}$ & $\mathbf{27.80}$\\ \hline 
\multicolumn{2}{|c|}{\textbf{MorCode}} & \hlg{$23.60$} & $53.21$ & $40.63$ & \hlg{$99.17$} & $100.00$ & $100.00$ & \hlg{$10.77$} & $19.09$ & $11.86$ & \hlg{$32.93$} & $95.14$ & $90.06$ & \hlg{$17.67$} & $58.80$ & $33.46$ & \hlg{$\mathbf{4.08}$} & $\mathbf{3.64}$ & $\mathbf{1.21}$\\ 
\hline \hline 
\multicolumn{2}{|c|}{\textbf{Average}} & \hlg{$27.42$} & $61.13$ & $49.67$ & \hlg{$36.55$} & $67.22$ & $56.16$ & \hlg{$15.75$} & $30.24$ & $23.56$ & \hlg{$23.80$} & $95.16$ & $90.13$ & \hlg{$27.55$} & $64.89$ & $54.99$ & \hlg{$\mathbf{5.63}$} & $\mathbf{12.93}$ & $\mathbf{7.46}$\\ \hline 
\multicolumn{20}{l}{\hl{$^*$FM: FaceMorpher, OCV: OpenCV, SG: StyleGAN2, WM: Webmorph, BE@AE: BPCER@APCER }}
\end{tabular}}}\vspace{-4mm} 
\end{center}
\end{table*}

\subsection{Datasets}
\vspace{1mm}\noindent\textbf{Training data.} \hl{In our experiments, we follow the protocols used in some recent MAD papers~\cite{caldeira_IDistill_EUSIPCO_2023, SYN_MAD_2022} and train SelfMAD on SMDD~\cite{Damer_SMDD_2022}. SMDD is a large-scale synthetic dataset specifically developed for face morphing attack detection. The dataset comprises high-quality bona fide and morphed samples, generated with a privacy-preserving approach based on StyleGAN2, ensuring diversity by simulating a wide range of facial features and morphing artifacts while avoiding real-world identity data. The dataset comprises $80,000$ images of size $1024 \times 1024$ pixels, including $50,000$ bona fide samples and $30,000$ morphing attacks, evenly split into training and testing subsets. For training SelfMAD, we use only the training subset, which contains $25,000$ bona fide images. Raw images are preprocessed by detecting faces in each image using Dlib~\cite{dlib_2009}. The detected square face regions are enlarged by a randomly selected margin between $4\%$ and $20\%$, then cropped and resized to $384 \times 384$ pixels. Additionally, segmentations of individual facial parts are generated for each cropped face using a pretrained SegFormer~\cite{xie_segformer_NIPS_2021}.} Both, the cropped RGB images and their corresponding segmentations, are then fed into SelfMAD and processed according to the steps described in Section~\ref{sec:methodology}.


In the experiments, we compare SelfMAD against both, unsupervised and discriminatively trained MAD methods. During the training of the unsupervised methods, we follow the protocols established by their respective authors and use the bona fide training data reported in the original papers. Similarly, the discriminative approaches are trained on the data utilized in their corresponding related works. Discriminatively trained MADs utilize bona fide images and morphing attacks from three publicly available datasets: LMA-DRD~\cite{Naser_PW_MAD_2021}, MorGAN~\cite{Damer_MorGAN_BTAS_2018}, and SMDD~\cite{Damer_SMDD_2022}. The LMA-DRD dataset includes morphs generated using OpenCV, with digital morphs labeled as "D" and re-digitized morphs (printed and scanned) labeled as "PS." The MorGAN dataset contains two types of morphs: LMA morphs created with OpenCV and GAN morphs produced using a DCGAN-based model. In the synthetic StyleGAN2-based SMDD dataset, both bona fide and morphing attack training subsets are utilized during training.


\vspace{1mm}\noindent\textbf{Testing data.} \hl{All MAD methods are tested on six standard datasets, representing various face morphing generation techniques: FRGC-Morphs~\cite{Sarkar2020_morphed_data}, FERET-Morphs~\cite{Sarkar2020_morphed_data}, FRLL-Morphs~\cite{Sarkar2020_morphed_data}, 
Morph-PIPE~\cite{Zhang_PIPE_IJCB_2024}, Greedy-DiM~\cite{Blasingame_greedy_IJCB_2024}, and MorCode~\cite{MorCode_BMVCW_2024}. FRGC-Morphs, FERET-Morphs, FRLL-Morphs, derived from respective source datasets FRGC~\cite{Phillips_FRGC}, FERET~\cite{PHILLIPS_FERET} and FRLL~\cite{debruine_jones_2017}, contain face morphing images generated by conventional landmark-based methods AMSL~\cite{Neubert_FRLL}, FaceMorpher
, OpenCV
, and WebMorph. Additionally, they also contain a deep--learning subset generated with StyleGAN2. Similarly, MorCode features GAN-based images generated using a VQ-GAN-based generator, with spherical interpolation in its latent space to create a single morphed image from two input face identities. In contrast, Morph-PIPE 
and Greedy-DiM 
are based on the latest generation of generative methods, i.e., diffusion models. While Greedy-DiM is derived from FRLL, MorCode, and Morph-PIPE morphs both originate from FRGC, and fulfill the quality constraints laid down by the International Civil Aviation Organization (ICAO).} Selected samples representing bona fide images and different morphing attacks are presented in Fig.~\ref{fig:examples}. Faces in all testing images were detected by Dlib~\cite{dlib_2009} and cropped out with a fixed margin of $12.5\%$.

\subsection{Evaluation metrics}
\hl{The model evaluation follows the testing protocol proposed in existing studies~\cite{Tapia_AlphaNet_WIFS_2023, Damer2022_OC_MAD_SPA} to enable consistent benchmarking and comparability. To ensure compliance with the ISO/IEC 30107-3~\cite{ISO301073}, we report the detection Equal Error Rate (EER), where the Attack Presentation Classification Error Rate (APCER), equals the Bona fide Presentation Classification Error Rate (BPCER). APCER here quantifies the proportion of attack samples misclassified as bona fide, while BPCER represents the proportion of bona fide samples misclassified as attacks. Besides EER, we also report the MAD performance at two operational points, i.e. BPCER at fixed APCER values of 5\% and 10\%.}

\begin{table*}[!t!]
\begin{center}
\caption{\hl{\textbf{Comparison of SelfMAD against discriminative MAD models} in terms of EER, BPCER@APCER (5\%) and BPCER@APCER (10\%). SelfMAD significantly outperforms competing methods across all morphing attack categories, except for FRGC and FERET StyleGAN2 (SG) morphs, where our method shows a slight underperformance.}} \label{tab:discriminative}
\resizebox{\linewidth}{!}{%
\hl{
\begin{tabular}{|l c c c c|c|c|c|c|c|c|c|c|c|c|c|c|c|c|c|c|c|}
\hline
\multicolumn{2}{|c|}{\multirow{2}{*}{\textbf{Model}}} & \multicolumn{1}{c|}{\textbf{Train}} &\multicolumn{2}{c|}{\multirow{2}{*}{\textbf{Test data:}}} & \multicolumn{3}{c|}{\textbf{FRGC-M}} & \multicolumn{3}{c|}{\textbf{FERET-M}} & \multicolumn{5}{c|}{\textbf{FRLL-M}} & \textbf{Morph-} & \textbf{Greedy-} & \multirow{2}{*}{\textbf{MorCode}} & \multirow{2}{*}{\textbf{Average}} \\ \cline{6-16} 
\multicolumn{2}{|c|}{} & \multicolumn{1}{c|}{\textbf{data}} & & & FM & OCV & SG & FM & OCV & SG & AMSL & FM & OCV & SG & WM & \textbf{PIPE} & \textbf{DiM} & & \\ 
\hline\hline
\multirow{15}{*}{\rotatebox{90}{\textbf{}}} & \multicolumn{1}{|c|}{\multirow{15}{*}{\rotatebox{90}{\textbf{MixFaceNet}~\cite{Boutros_MixFaceNet_2021}}}} & \multirow{3}{*}{D} & \multicolumn{2}{|c|}{\hlg{EER}} & \hlg{$32.43$} & \hlg{$20.34$} & \hlg{$19.59$} & \hlg{$27.92$} & \hlg{$33.59$} & \hlg{$37.49$} & \hlg{$23.15$} & \hlg{$10.84$} & \hlg{$10.84$} & \hlg{$19.70$} & \hlg{$31.03$} & \hlg{$46.64$} & \hlg{$45.10$} & \hlg{$39.83$} & \hlg{$28.46$}\\ \cline{4-20} 
 & \multicolumn{1}{|c|}{} & & \multicolumn{1}{|c|}{\multirow{2}{*}{\rotatebox{0}{BE@AE(\%)}}} & $5.00$ & $85.89$ & $51.97$ & $64.63$ & $69.38$ & $79.40$ & $91.87$ & $71.40$ & $23.71$ & $39.97$ & $54.58$ & $85.26$ & $98.41$ & $93.40$ & $92.79$ & $71.62$\\ \cline{5-20} 
 & \multicolumn{1}{|c|}{} & & \multicolumn{1}{|c|}{} & $10.00$ & $73.44$ & $37.03$ & $41.91$ & $57.09$ & $68.43$ & $81.10$ & $61.47$ & $13.66$ & $18.02$ & $39.20$ & $71.01$ & $95.75$ & $85.60$ & $84.90$ & $59.19$\\ \cline{3-20} 
 & \multicolumn{1}{|c|}{} & \multirow{3}{*}{PS} & \multicolumn{2}{|c|}{\hlg{EER}} & \hlg{$25.54$} & \hlg{$30.13$} & \hlg{$19.75$} & \hlg{$31.40$} & \hlg{$29.62$} & \hlg{$30.19$} & \hlg{$19.70$} & \hlg{$8.87$} & \hlg{$9.85$} & \hlg{$11.82$} & \hlg{$30.05$} & \hlg{$61.51$} & \hlg{$41.67$} & \hlg{$43.84$} & \hlg{$28.14$}\\ \cline{4-20} 
 & \multicolumn{1}{|c|}{} & & \multicolumn{1}{|c|}{\multirow{2}{*}{\rotatebox{0}{BE@AE(\%)}}} & $5.00$ & $75.31$ & $79.25$ & $46.89$ & $82.61$ & $77.13$ & $83.93$ & $56.55$ & $15.38$ & $28.26$ & $28.07$ & $76.82$ & $100.00$ & $89.00$ & $92.19$ & $66.53$\\ \cline{5-20} 
 & \multicolumn{1}{|c|}{} & & \multicolumn{1}{|c|}{} & $10.00$ & $55.19$ & $64.83$ & $33.82$ & $73.53$ & $66.16$ & $68.43$ & $45.06$ & $8.33$ & $14.00$ & $17.76$ & $65.36$ & $99.62$ & $78.80$ & $84.45$ & $55.38$\\ \cline{3-20} 
 & \multicolumn{1}{|c|}{} & \multirow{3}{*}{LMA} & \multicolumn{2}{|c|}{\hlg{EER}} & \hlg{$35.76$} & \hlg{$32.33$} & \hlg{$17.26$} & \hlg{$36.07$} & \hlg{$33.95$} & \hlg{$27.43$} & \hlg{$27.09$} & \hlg{$27.09$} & \hlg{$18.23$} & \hlg{$29.06$} & \hlg{$45.81$} & \hlg{$35.18$} & \hlg{$40.69$} & \hlg{$37.28$} & \hlg{$31.66$}\\ \cline{4-20} 
 & \multicolumn{1}{|c|}{} & & \multicolumn{1}{|c|}{\multirow{2}{*}{\rotatebox{0}{BE@AE(\%)}}} & $5.00$ & $85.58$ & $85.06$ & $49.38$ & $84.12$ & $89.41$ & $76.37$ & $84.64$ & $69.93$ & $44.64$ & $83.39$ & $94.35$ & $78.15$ & $93.40$ & $83.84$ & $78.73$\\ \cline{5-20} 
 & \multicolumn{1}{|c|}{} & & \multicolumn{1}{|c|}{} & $10.00$ & $76.35$ & $72.93$ & $30.81$ & $76.75$ & $78.45$ & $61.63$ & $74.53$ & $49.05$ & $27.27$ & $71.11$ & $90.01$ & $66.16$ & $88.00$ & $74.51$ & $66.97$\\ \cline{3-20} 
 & \multicolumn{1}{|c|}{} & \multirow{3}{*}{GAN} & \multicolumn{2}{|c|}{\hlg{EER}} & \hlg{$26.25$} & \hlg{$29.91$} & \hlg{$33.40$} & \hlg{$44.08$} & \hlg{$44.72$} & \hlg{$48.19$} & \hlg{$45.32$} & \hlg{$46.80$} & \hlg{$39.41$} & \hlg{$39.41$} & \hlg{$44.83$} & \hlg{$56.62$} & \hlg{$48.04$} & \hlg{$54.42$} & \hlg{$42.96$}\\ \cline{4-20} 
 & \multicolumn{1}{|c|}{} & & \multicolumn{1}{|c|}{\multirow{2}{*}{\rotatebox{0}{BE@AE(\%)}}} & $5.00$ & $76.45$ & $80.71$ & $89.94$ & $95.46$ & $90.55$ & $94.90$ & $95.45$ & $97.08$ & $96.56$ & $85.76$ & $95.25$ & $99.85$ & $97.00$ & $99.39$ & $92.45$\\ \cline{5-20} 
 & \multicolumn{1}{|c|}{} & & \multicolumn{1}{|c|}{} & $10.00$ & $60.89$ & $68.78$ & $79.36$ & $90.74$ & $83.55$ & $88.66$ & $88.83$ & $92.35$ & $93.94$ & $78.97$ & $92.55$ & $99.24$ & $93.40$ & $95.75$ & $86.21$\\ \cline{3-20} 
 & \multicolumn{1}{|c|}{} & \multirow{3}{*}{SMDD} & \multicolumn{2}{|c|}{\hlg{EER}} & \hlg{$14.32$} & \hlg{$14.35$} & \hlg{$\mathbf{13.03}$} & \hlg{$19.99$} & \hlg{$20.62$} & \hlg{$36.64$} & \hlg{$31.03$} & \hlg{$8.37$} & \hlg{$9.85$} & \hlg{$38.92$} & \hlg{$31.03$} & \hlg{$33.69$} & \hlg{$39.71$} & \hlg{$37.55$} & \hlg{$24.94$}\\ \cline{4-20} 
 & \multicolumn{1}{|c|}{} & & \multicolumn{1}{|c|}{\multirow{2}{*}{\rotatebox{0}{BE@AE(\%)}}} & $5.00$ & $33.40$ & $35.27$ & $\mathbf{26.97}$ & $53.69$ & $53.88$ & $87.33$ & $65.56$ & $11.51$ & $15.89$ & $84.94$ & $78.38$ & $85.66$ & $90.20$ & $86.12$ & $57.77$\\ \cline{5-20} 
 & \multicolumn{1}{|c|}{} & & \multicolumn{1}{|c|}{} & $10.00$ & $20.95$ & $20.95$ & $\mathbf{15.87}$ & $34.97$ & $39.51$ & $75.24$ & $60.00$ & $8.16$ & $10.07$ & $76.27$ & $67.16$ & $73.75$ & $86.20$ & $76.10$ & $47.51$\\ \cline{2-20} 
\multirow{15}{*}{\rotatebox{90}{\textbf{Discriminativelly trained}}} & \multicolumn{1}{|c|}{\multirow{15}{*}{\rotatebox{90}{\textbf{PW-MAD}~\cite{Naser_PW_MAD_2021}}}} & \multirow{3}{*}{D} & \multicolumn{2}{|c|}{\hlg{EER}} & \hlg{$36.44$} & \hlg{$38.64$} & \hlg{$36.70$} & \hlg{$14.60$} & \hlg{$11.55$} & \hlg{$40.61$} & \hlg{$35.96$} & \hlg{$14.78$} & \hlg{$11.33$} & \hlg{$35.96$} & \hlg{$35.96$} & \hlg{$13.36$} & \hlg{$17.16$} & \hlg{$17.87$} & \hlg{$25.78$}\\ \cline{4-20} 
 & \multicolumn{1}{|c|}{} & & \multicolumn{1}{|c|}{\multirow{2}{*}{\rotatebox{0}{BE@AE(\%)}}} & $5.00$ & $95.12$ & $91.29$ & $97.82$ & $47.26$ & $21.55$ & $95.84$ & $87.08$ & $69.85$ & $22.19$ & $93.94$ & $90.09$ & $54.93$ & $64.20$ & $50.15$ & $70.09$\\ \cline{5-20} 
 & \multicolumn{1}{|c|}{} & & \multicolumn{1}{|c|}{} & $10.00$ & $86.00$ & $83.51$ & $93.67$ & $25.33$ & $13.99$ & $88.09$ & $71.95$ & $33.51$ & $12.04$ & $80.61$ & $75.68$ & $22.76$ & $36.60$ & $33.92$ & $54.12$\\ \cline{3-20} 
 & \multicolumn{1}{|c|}{} & \multirow{3}{*}{PS} & \multicolumn{2}{|c|}{\hlg{EER}} & \hlg{$42.81$} & \hlg{$53.80$} & \hlg{$38.02$} & \hlg{$14.88$} & \hlg{$11.48$} & \hlg{$45.29$} & \hlg{$22.66$} & \hlg{$15.76$} & \hlg{$7.88$} & \hlg{$28.57$} & \hlg{$26.60$} & \hlg{$20.97$} & \hlg{$33.82$} & \hlg{$34.74$} & \hlg{$28.38$}\\ \cline{4-20} 
 & \multicolumn{1}{|c|}{} & & \multicolumn{1}{|c|}{\multirow{2}{*}{\rotatebox{0}{BE@AE(\%)}}} & $5.00$ & $94.92$ & $98.86$ & $66.29$ & $52.93$ & $17.39$ & $97.16$ & $53.75$ & $34.28$ & $12.29$ & $78.15$ & $66.67$ & $65.63$ & $83.40$ & $85.13$ & $64.77$\\ \cline{5-20} 
 & \multicolumn{1}{|c|}{} & & \multicolumn{1}{|c|}{} & $10.00$ & $89.52$ & $97.20$ & $59.13$ & $32.70$ & $12.29$ & $93.95$ & $46.85$ & $22.42$ & $6.72$ & $67.02$ & $48.81$ & $50.30$ & $70.00$ & $73.60$ & $55.04$\\ \cline{3-20} 
 & \multicolumn{1}{|c|}{} & \multirow{3}{*}{LMA} & \multicolumn{2}{|c|}{\hlg{EER}} & \hlg{$21.53$} & \hlg{$12.93$} & \hlg{$19.08$} & \hlg{$18.50$} & \hlg{$24.66$} & \hlg{$26.44$} & \hlg{$18.23$} & \hlg{$10.34$} & \hlg{$12.81$} & \hlg{$19.21$} & \hlg{$39.41$} & \hlg{$17.27$} & \hlg{$17.16$} & \hlg{$21.15$} & \hlg{$19.91$}\\ \cline{4-20} 
 & \multicolumn{1}{|c|}{} & & \multicolumn{1}{|c|}{\multirow{2}{*}{\rotatebox{0}{BE@AE(\%)}}} & $5.00$ & $93.67$ & $48.86$ & $61.00$ & $75.99$ & $75.99$ & $77.69$ & $47.49$ & $26.72$ & $27.35$ & $42.39$ & $92.22$ & $54.32$ & $40.80$ & $56.68$ & $58.66$\\ \cline{5-20} 
 & \multicolumn{1}{|c|}{} & & \multicolumn{1}{|c|}{} & $10.00$ & $66.29$ & $20.12$ & $39.21$ & $48.58$ & $65.97$ & $60.87$ & $33.47$ & $12.54$ & $19.66$ & $30.20$ & $83.70$ & $33.69$ & $28.60$ & $39.98$ & $41.63$\\ \cline{3-20} 
 & \multicolumn{1}{|c|}{} & \multirow{3}{*}{GAN} & \multicolumn{2}{|c|}{\hlg{EER}} & \hlg{$47.17$} & \hlg{$46.01$} & \hlg{$24.09$} & \hlg{$12.19$} & \hlg{$25.02$} & \hlg{$43.23$} & \hlg{$47.78$} & \hlg{$21.18$} & \hlg{$31.03$} & \hlg{$48.28$} & \hlg{$50.25$} & \hlg{$32.18$} & \hlg{$15.20$} & \hlg{$54.89$} & \hlg{$35.61$}\\ \cline{4-20} 
 & \multicolumn{1}{|c|}{} & & \multicolumn{1}{|c|}{\multirow{2}{*}{\rotatebox{0}{BE@AE(\%)}}} & $5.00$ & $99.79$ & $99.07$ & $100.00$ & $31.95$ & $70.51$ & $88.66$ & $92.41$ & $66.07$ & $97.22$ & $97.46$ & $93.61$ & $96.28$ & $39.60$ & $87.71$ & $82.88$\\ \cline{5-20} 
 & \multicolumn{1}{|c|}{} & & \multicolumn{1}{|c|}{} & $10.00$ & $98.76$ & $97.72$ & $100.00$ & $16.82$ & $54.63$ & $80.15$ & $88.00$ & $48.80$ & $83.21$ & $90.59$ & $89.76$ & $81.49$ & $22.20$ & $83.16$ & $73.95$\\ \cline{3-20} 
 & \multicolumn{1}{|c|}{} & \multirow{3}{*}{SMDD} & \multicolumn{2}{|c|}{\hlg{EER}} & \hlg{$15.97$} & \hlg{$19.33$} & \hlg{$15.36$} & \hlg{$8.36$} & \hlg{$16.09$} & \hlg{$20.91$} & \hlg{$4.43$} & \hlg{$1.97$} & \hlg{$2.46$} & \hlg{$17.24$} & \hlg{$9.85$} & \hlg{$20.40$} & \hlg{$42.16$} & \hlg{$39.29$} & \hlg{$16.70$}\\ \cline{4-20} 
 & \multicolumn{1}{|c|}{} & & \multicolumn{1}{|c|}{\multirow{2}{*}{\rotatebox{0}{BE@AE(\%)}}} & $5.00$ & $51.76$ & $56.85$ & $35.68$ & $42.34$ & $74.48$ & $81.47$ & $4.64$ & $1.72$ & $1.80$ & $34.62$ & $12.29$ & $55.39$ & $95.40$ & $93.63$ & $45.86$\\ \cline{5-20} 
 & \multicolumn{1}{|c|}{} & & \multicolumn{1}{|c|}{} & $10.00$ & $29.25$ & $38.90$ & $21.06$ & $5.10$ & $35.92$ & $53.31$ & $2.21$ & $1.29$ & $0.90$ & $26.43$ & $9.75$ & $37.78$ & $86.80$ & $86.65$ & $31.10$\\ \cline{2-20}
\multirow{15}{*}{\rotatebox{90}{\textbf{}}} & \multicolumn{1}{|c|}{\multirow{15}{*}{\rotatebox{90}{\textbf{Inception}~\cite{Ramachandra_Inception_2020}}}} & \multirow{3}{*}{D} & \multicolumn{2}{|c|}{\hlg{EER}} & \hlg{$42.22$} & \hlg{$30.26$} & \hlg{$39.19$} & \hlg{$14.74$} & \hlg{$15.59$} & \hlg{$\mathbf{11.27}$} & \hlg{$55.67$} & \hlg{$39.41$} & \hlg{$28.57$} & \hlg{$25.62$} & \hlg{$65.52$} & \hlg{$33.63$} & \hlg{$31.86$} & \hlg{$10.91$} & \hlg{$31.75$}\\ \cline{4-20} 
 & \multicolumn{1}{|c|}{} & & \multicolumn{1}{|c|}{\multirow{2}{*}{\rotatebox{0}{BE@AE(\%)}}} & $5.00$ & $96.89$ & $93.57$ & $89.00$ & $31.57$ & $32.33$ & $\mathbf{21.93}$ & $99.26$ & $81.36$ & $69.45$ & $67.59$ & $99.92$ & $95.90$ & $85.40$ & $21.70$ & $70.42$\\ \cline{5-20} 
 & \multicolumn{1}{|c|}{} & & \multicolumn{1}{|c|}{} & $10.00$ & $90.15$ & $80.29$ & $78.11$ & $21.55$ & $20.98$ & $\mathbf{13.04}$ & $98.48$ & $70.45$ & $57.82$ & $57.77$ & $99.75$ & $82.09$ & $77.40$ & $12.06$ & $61.43$\\ \cline{3-20} 
 & \multicolumn{1}{|c|}{} & \multirow{3}{*}{PS} & \multicolumn{2}{|c|}{\hlg{EER}} & \hlg{$43.87$} & \hlg{$20.50$} & \hlg{$24.86$} & \hlg{$30.33$} & \hlg{$27.64$} & \hlg{$16.65$} & \hlg{$50.25$} & \hlg{$32.02$} & \hlg{$41.38$} & \hlg{$33.50$} & \hlg{$39.90$} & \hlg{$8.30$} & \hlg{$51.96$} & \hlg{$27.58$} & \hlg{$32.05$}\\ \cline{4-20} 
 & \multicolumn{1}{|c|}{} & & \multicolumn{1}{|c|}{\multirow{2}{*}{\rotatebox{0}{BE@AE(\%)}}} & $5.00$ & $93.36$ & $53.11$ & $47.10$ & $69.57$ & $64.84$ & $31.38$ & $99.49$ & $78.95$ & $92.79$ & $75.04$ & $93.28$ & $20.18$ & $95.20$ & $72.00$ & $70.45$\\ \cline{5-20} 
 & \multicolumn{1}{|c|}{} & & \multicolumn{1}{|c|}{} & $10.00$ & $85.89$ & $37.34$ & $38.28$ & $57.84$ & $53.69$ & $22.12$ & $98.16$ & $68.99$ & $88.94$ & $68.33$ & $87.96$ & $5.16$ & $92.20$ & $61.08$ & $61.86$\\ \cline{3-20} 
 & \multicolumn{1}{|c|}{} & \multirow{3}{*}{LMA} & \multicolumn{2}{|c|}{\hlg{EER}} & \hlg{$27.19$} & \hlg{$30.26$} & \hlg{$42.39$} & \hlg{$27.21$} & \hlg{$36.57$} & \hlg{$29.91$} & \hlg{$12.81$} & \hlg{$8.37$} & \hlg{$19.21$} & \hlg{$19.70$} & \hlg{$24.14$} & \hlg{$58.38$} & \hlg{$25.98$} & \hlg{$29.59$} & \hlg{$27.98$}\\ \cline{4-20} 
 & \multicolumn{1}{|c|}{} & & \multicolumn{1}{|c|}{\multirow{2}{*}{\rotatebox{0}{BE@AE(\%)}}} & $5.00$ & $79.46$ & $89.00$ & $91.70$ & $70.32$ & $86.77$ & $81.85$ & $35.26$ & $20.10$ & $63.47$ & $66.94$ & $72.73$ & $100.00$ & $66.00$ & $63.73$ & $70.52$\\ \cline{5-20} 
 & \multicolumn{1}{|c|}{} & & \multicolumn{1}{|c|}{} & $10.00$ & $60.48$ & $76.04$ & $84.34$ & $57.47$ & $73.72$ & $68.81$ & $20.37$ & $5.76$ & $41.52$ & $48.61$ & $64.13$ & $99.92$ & $46.00$ & $51.75$ & $57.06$\\ \cline{3-20} 
 & \multicolumn{1}{|c|}{} & \multirow{3}{*}{GAN} & \multicolumn{2}{|c|}{\hlg{EER}} & \hlg{$49.76$} & \hlg{$49.08$} & \hlg{$41.67$} & \hlg{$51.10$} & \hlg{$52.73$} & \hlg{$35.79$} & \hlg{$75.86$} & \hlg{$67.49$} & \hlg{$59.61$} & \hlg{$67.49$} & \hlg{$59.61$} & \hlg{$40.76$} & \hlg{$29.90$} & \hlg{$61.98$} & \hlg{$53.06$}\\ \cline{4-20} 
 & \multicolumn{1}{|c|}{} & & \multicolumn{1}{|c|}{\multirow{2}{*}{\rotatebox{0}{BE@AE(\%)}}} & $5.00$ & $99.69$ & $98.76$ & $95.23$ & $95.65$ & $98.11$ & $82.99$ & $100.00$ & $99.31$ & $98.94$ & $98.53$ & $99.84$ & $95.45$ & $86.60$ & $100.00$ & $96.36$\\ \cline{5-20} 
 & \multicolumn{1}{|c|}{} & & \multicolumn{1}{|c|}{} & $10.00$ & $98.65$ & $96.58$ & $89.83$ & $92.25$ & $96.98$ & $72.40$ & $100.00$ & $98.71$ & $96.97$ & $97.55$ & $99.43$ & $89.98$ & $73.00$ & $99.70$ & $93.00$\\ \cline{3-20} 
 & \multicolumn{1}{|c|}{} & \multirow{3}{*}{SMDD} & \multicolumn{2}{|c|}{\hlg{EER}} & \hlg{$18.82$} & \hlg{$23.38$} & \hlg{$32.82$} & \hlg{$19.42$} & \hlg{$43.80$} & \hlg{$38.98$} & \hlg{$7.88$} & \hlg{$2.96$} & \hlg{$15.27$} & \hlg{$19.70$} & \hlg{$11.33$} & \hlg{$11.90$} & \hlg{$56.86$} & \hlg{$11.24$} & \hlg{$22.45$}\\ \cline{4-20} 
 & \multicolumn{1}{|c|}{} & & \multicolumn{1}{|c|}{\multirow{2}{*}{\rotatebox{0}{BE@AE(\%)}}} & $5.00$ & $54.15$ & $67.12$ & $99.90$ & $69.57$ & $100.00$ & $99.43$ & $11.03$ & $2.66$ & $37.59$ & $45.58$ & $18.18$ & $26.02$ & $100.00$ & $21.55$ & $53.77$\\ \cline{5-20} 
 & \multicolumn{1}{|c|}{} & & \multicolumn{1}{|c|}{} & $10.00$ & $35.68$ & $47.61$ & $97.20$ & $46.12$ & $99.43$ & $93.76$ & $7.17$ & $2.23$ & $22.60$ & $35.84$ & $12.69$ & $14.64$ & $99.80$ & $13.81$ & $44.90$\\ \hline \hline 
\multicolumn{1}{|c}{} & & & \multicolumn{2}{|c|}{\hlg{EER}} & \hlg{$\mathbf{5.59}$} & \hlg{$\mathbf{2.59}$} & \hlg{$15.84$} & \hlg{$\mathbf{3.19}$} & \hlg{$\mathbf{1.13}$} & \hlg{$18.14$} & \hlg{$\mathbf{0.99}$} & \hlg{$\mathbf{0.00}$} & \hlg{$\mathbf{0.00}$} & \hlg{$\mathbf{10.34}$} & \hlg{$\mathbf{3.45}$} & \hlg{$\mathbf{5.89}$} & \hlg{$\mathbf{7.60}$} & \hlg{$\mathbf{4.08}$} & \hlg{$\mathbf{5.63}$}\\ \cline{4-20} 
 \multicolumn{3}{|c}{\textbf{SelfMAD}} & \multicolumn{1}{|c|}{\multirow{2}{*}{\rotatebox{0}{BE@AE(\%)}}} & $5.00$ & $\mathbf{6.43}$ & $\mathbf{1.14}$ & $45.23$ & $\mathbf{1.70}$ & $\mathbf{0.57}$ & $46.12$ & $\mathbf{0.05}$ & $\mathbf{0.26}$ & $\mathbf{0.00}$ & $\mathbf{24.22}$ & $\mathbf{1.64}$ & $\mathbf{12.44}$ & $\mathbf{37.60}$ & $\mathbf{3.64}$ & $\mathbf{12.93}$\\ \cline{5-20} 
 & & & \multicolumn{1}{|c|}{} & $10.00$ & $\mathbf{2.80}$ & $\mathbf{0.41}$ & $25.52$ & $\mathbf{0.38}$ & $\mathbf{0.38}$ & $32.33$ & $\mathbf{0.05}$ & $\mathbf{0.17}$ & $\mathbf{0.00}$ & $\mathbf{12.52}$ & $\mathbf{0.41}$ & $\mathbf{0.53}$ & $\mathbf{27.80}$ & $\mathbf{1.21}$ & $\mathbf{7.46}$\\ \hline 
 \multicolumn{20}{l}{$^*$FM: FaceMorpher, OCV: OpenCV, SG: StyleGAN2, WM: Webmorph, BE@AE: BPCER@APCER} \\
\end{tabular}}}\vspace{-4mm}
\end{center}
\end{table*}

\subsection{Implementation details}
The input to SelfMAD represents RGB bona fide face images. Face blending masks are generated by combining face segmentations created with a SegFormer face parser, pretrained on the CelebAMask-HQ dataset\footnote{https://huggingface.co/jonathandinu/face-parsing}. Once SelfMAD generates augmented face images with superimposed pixel and frequency artifacts, bona fide samples along with simulated morphing attacks are fed into a binary classifier. For the classifier we consider $4$ different CNN architectures: EfficientNet-B4, ResNet-152, Swin-B and HRNet-W18. All of them were pretrained on ImageNet and finetuned until convergence on samples generated during earlier stages of SelfMAD. \hl{In this paper, we only report results obtained with HRNet as a backbone, as it achieved the best overall performance on considered testing datasets. A comparison of HRNet's performance against other backbones is provided in the supplementary material.} The classifier is optimized by SGD combined with SAM~\cite{foret_optim_SAM_ICLR_2021}. The radius $\rho$ of SAM is set to $0.05$ with a momentum of $0.9$, while the learning rate equals $0.001$. SelfMAD is implemented in Python 3.10 with PyTorch 2.4 and CUDA 12.5. Experiments were run on NVIDIA GeForce RTX 4090. The source code of SelfMAD is available at \href{https://github.com/LeonTodorov/SelfMAD}{https://github.com/LeonTodorov/SelfMAD}.

\begin{table*}[!tb!]
\begin{center}
\caption{\hl{\textbf{Ablation study evaluating different SelfMAD components.} While the Pixel Artifact Generator (PAG) alone achieves strong morphing detection performance, combining it with the Frequency Artifact Generator (FAG) enhances the overall SelfMAD performance across various attack categories.}} \label{tab:ablation}
\resizebox{\linewidth}{!}{%
\hl{
\begin{tabular}{|l c c|c|c|c|c|c|c|c|c|c|c|c|c|c|c|c|c|c|}
\hline
\multicolumn{3}{|c|}{\multirow{2}{*}{\textbf{Test data:}}} & \multicolumn{3}{c|}{\textbf{FRGC-M}} & \multicolumn{3}{c|}{\textbf{FERET-M}} & \multicolumn{5}{c|}{\textbf{FRLL-M}} & \textbf{Morph-} & \textbf{Greedy-} & \multirow{2}{*}{\textbf{MorCode}} & \multirow{2}{*}{\textbf{Average}} \\ \cline{4-14} 
& & & FM & OCV & SG & FM & OCV & SG & AMSL & FM & OCV & SG & WM & \textbf{PIPE} & \textbf{DiM} & & \\ 
\hline\hline
\multirow{3}{*}{\rotatebox{0}{\textbf{PAG}}} & \multicolumn{2}{|c|}{\hlg{EER}} & \hlg{$7.24$} & \hlg{$3.52$} & \hlg{$\mathbf{8.02}$} & \hlg{$6.17$} & \hlg{$2.83$} & \hlg{$21.05$} & \hlg{$1.48$} & \hlg{$\mathbf{0.00}$} & \hlg{$\mathbf{0.00}$} & \hlg{$\mathbf{9.36}$} & \hlg{$8.37$} & \hlg{$\mathbf{2.21}$} & \hlg{$26.00$} & \hlg{$\mathbf{3.08}$} & \hlg{$7.10$}\\ \cline{2-18} 
 & \multicolumn{1}{|c}{\multirow{2}{*}{\rotatebox{0}{BE@AE(\%)}}} & \multicolumn{1}{|c|}{$5.00$} & $11.31$ & $2.39$ & $\mathbf{18.88}$ & $7.37$ & $1.51$ & $48.02$ & $0.28$ & $\mathbf{0.26}$ & $\mathbf{0.00}$ & $\mathbf{13.09}$ & $12.37$ & $\mathbf{0.08}$ & $95.40$ & $\mathbf{1.59}$ & $15.18$\\ \cline{3-18} 
 & \multicolumn{1}{|c}{} & \multicolumn{1}{|c|}{$10.00$} & $4.56$ & $0.52$ & $\mathbf{6.33}$ & $2.46$ & $1.32$ & $36.29$ & $0.09$ & $0.26$ & $\mathbf{0.00}$ & $\mathbf{9.00}$ & $5.57$ & $\mathbf{0.08}$ & $90.20$ & $\mathbf{0.38}$ & $11.22$\\ \hline 
\multirow{3}{*}{\rotatebox{0}{\textbf{FAG}}} & \multicolumn{2}{|c|}{\hlg{EER}} & \hlg{$66.41$} & \hlg{$47.88$} & \hlg{$36.05$} & \hlg{$54.36$} & \hlg{$54.78$} & \hlg{$23.39$} & \hlg{$83.25$} & \hlg{$82.27$} & \hlg{$73.40$} & \hlg{$46.80$} & \hlg{$80.79$} & \hlg{$97.93$} & \hlg{$78.40$} & \hlg{$97.79$} & \hlg{$65.96$}\\ \cline{2-18} 
 & \multicolumn{1}{|c}{\multirow{2}{*}{\rotatebox{0}{BE@AE(\%)}}} & \multicolumn{1}{|c|}{$5.00$} & $100.00$ & $100.00$ & $98.13$ & $98.87$ & $98.30$ & $48.58$ & $100.00$ & $100.00$ & $100.00$ & $100.00$ & $100.00$ & $100.00$ & $100.00$ & $100.00$ & $95.99$\\ \cline{3-18} 
 & \multicolumn{1}{|c}{} & \multicolumn{1}{|c|}{$10.00$} & $100.00$ & $99.59$ & $94.50$ & $95.84$ & $95.65$ & $37.43$ & $100.00$ & $100.00$ & $100.00$ & $99.67$ & $100.00$ & $100.00$ & $100.00$ & $100.00$ & $94.48$\\ \hline 
\multirow{3}{*}{\rotatebox{0}{\textbf{SelfMAD}}} & \multicolumn{2}{|c|}{\hlg{EER}} & \hlg{$\mathbf{5.59}$} & \hlg{$\mathbf{2.59}$} & \hlg{$15.84$} & \hlg{$\mathbf{3.19}$} & \hlg{$\mathbf{1.13}$} & \hlg{$\mathbf{18.14}$} & \hlg{$\mathbf{0.99}$} & \hlg{$\mathbf{0.00}$} & \hlg{$\mathbf{0.00}$} & \hlg{$10.34$} & \hlg{$\mathbf{3.45}$} & \hlg{$5.89$} & \hlg{$\mathbf{7.60}$} & \hlg{$4.08$} & \hlg{$\mathbf{5.63}$}\\ \cline{2-18} 
 & \multicolumn{1}{|c}{\multirow{2}{*}{\rotatebox{0}{BE@AE(\%)}}} & \multicolumn{1}{|c|}{$5.00$} & $\mathbf{6.43}$ & $\mathbf{1.14}$ & $45.23$ & $\mathbf{1.70}$ & $\mathbf{0.57}$ & $\mathbf{46.12}$ & $\mathbf{0.05}$ & $\mathbf{0.26}$ & $\mathbf{0.00}$ & $24.22$ & $\mathbf{1.64}$ & $12.44$ & $\mathbf{37.60}$ & $3.64$ & $\mathbf{12.93}$\\ \cline{3-18} 
 & \multicolumn{1}{|c}{} & \multicolumn{1}{|c|}{$10.00$} & $\mathbf{2.80}$ & $\mathbf{0.41}$ & $25.52$ & $\mathbf{0.38}$ & $\mathbf{0.38}$ & $\mathbf{32.33}$ & $\mathbf{0.05}$ & $\mathbf{0.17}$ & $\mathbf{0.00}$ & $12.52$ & $\mathbf{0.41}$ & $0.53$ & $\mathbf{27.80}$ & $1.21$ & $\mathbf{7.46}$\\ \hline 
\multicolumn{18}{l}{$^*$FM: FaceMorpher, OCV: OpenCV, SG: StyleGAN2, WM: Webmorph, BE@AE: BPCER@APCER} \\
\end{tabular}}}\vspace{-6mm}
\end{center}
\end{table*}

\vspace{-5pt}
\section{Results}\label{Sec: Results}
In this section, we first compare SelfMAD against competitive one-class and discriminatively trained MAD models. Next, we investigate the impact of different SelfMAD components on model's performance. Finally, we visualize the regions of interest identified by SelfMAD when the model classifies a sample as a morphing attack.

\begin{figure*}[!t!]
    \begin{center}
        \includegraphics[width=0.96\linewidth]{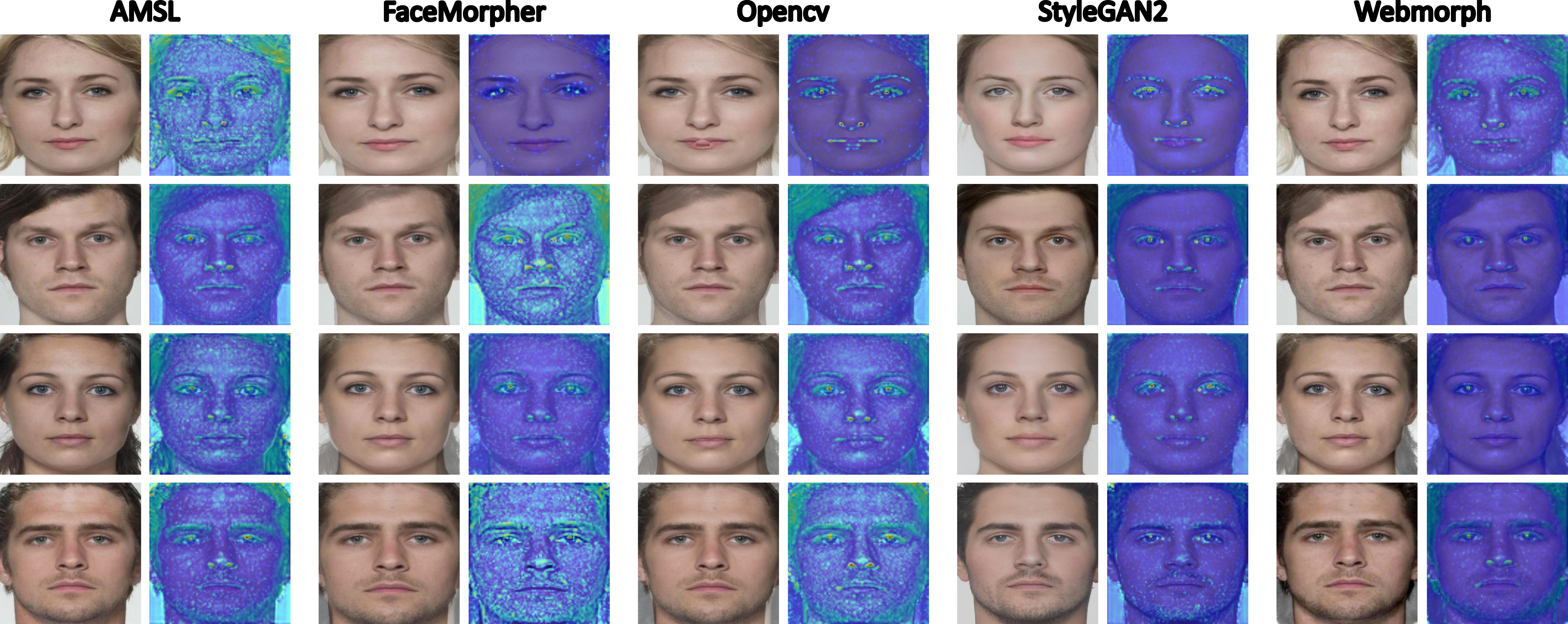}\vspace{-1mm}
        \caption{\textbf{GradCAM visualizations} of SelfMAD, generated with HRNet-W18 as backbone. Heatmaps highlight eyes, nostrils, lip borders and hair, as areas where morphing pixel irregularities are most prominent. Note that StyleGAN2 images are also prone to asymmetries and blurry edges in these regions.}\label{fig:heatmaps}
    \end{center}
\vspace{-6mm} \end{figure*} 

\vspace{1mm}\noindent\textbf{Comparison against one-class MADs.} SelfMAD is designed to be trained exclusively with bona fide input samples, so we first compare its performance against other highly competitive one-class methods. The results are summarized in Table~\ref{tab:unsupervised}. As shown, SelfMAD achieves the best overall performance across various face morphing attack types, \hl{surpassing the runner-up SPL-MAD~\cite{Damer2022_OC_MAD_SPA} by $10.12\%$ in terms of EER, and by $17.31\%$ and $16.1\%$ in terms of BPCER@APCER 5\% and 10\%, respectively. We note, that SelfMAD is especially successful in the detection of more recent, advanced GAN and diffusion-based morphing attacks from Morph-PIPE, Greedy-DiM and MorCode, improving the average EER performance of the runner-ups in these three datasets, SBI~\cite{Shiohara_SBI_CVPR_2022} and MAD-DDPM~\cite{ivanovska2023mad_ddpm}, by more than $78\%$. However, while SelfMAD outperforms other models in most morphing attack categories, it exhibits slight underperformance in detecting StyleGAN2 (SG) morphs. In this specific category, the MAD approach based on face image quality assessment, FIQA-MagFace~\cite{damer_iqa_IETB_2022}, demonstrates superior results despite having one of the lowest overall performance with an average EER of $27.42\%$. Notably, SBI, a self-supervised model, that is conceptually closest to SelfMAD, is outperformed by our model by a large margin of $21.92\%$ EER, $51.96\%$ BPCER@APCER $5\%$ and $47.53\%$ BPCER@APCER $10\%$.} This finding highlights the advantages of our proposed approach for the specific task of face morphing attack detection.

\vspace{1mm}\noindent\textbf{Comparison against discriminative MADs.} We also compare SelfMAD against state-of-the-art discriminative MAD techniques, i.e., MixFaceNet~\cite{Boutros_MixFaceNet_2021}, PW-MAD~\cite{Naser_PW_MAD_2021} and Inception~\cite{Ramachandra_Inception_2020}. This comparison aims to evaluate the generalizability of the discriminatively trained models, similar to the cross-dataset experiments conducted in~\cite{Damer2022_OC_MAD_SPA} and \cite{ivanovska2023mad_ddpm}. For this purpose, we train each competing model in a two-class setting, using training images from a different set of morphing attacks than those used in the evaluation phase. We run five training sessions, each time using a new, distinct type of face morphing attack as the training data and then test the trained models on testing splits from another datasets. \hl{The results, presented in Table~\ref{tab:discriminative}, show that SelfMAD is again the top-performer, surpassing the runner-up, PW-MAD trained on SMDD, by a large margin of $11.07\%$ in terms of EER, $32.93\%$ in terms of BPCER@APCER $5\%$ and $23.64\%$ in terms of BPCER@APCER $10\%$. However, we again note that SelfMAD shows a slight underperformance in FRGC and FERET StyleGANv2 (SG) morphs.}

\vspace{1mm}\noindent\textbf{Ablation.} To assess the contributions of individual components in SelfMAD, we conduct an ablation study, where we evaluate the model's detection performance under three scenarios: (1) we generate only pixel-space artifacts, (2) we generate only frequency-space artifacts, and (3) both, pixel and frequency artifacts, are added to the original bona fide input image. The results of this analysis, conducted with the best-performing SelfMAD backbone, HRNet-W18, are summarized in Table~\ref{tab:ablation}. Notably, the generation of pixel-space artifacts alone yields strong morphing attack detection performance across various morphing techniques, \hl{including some advanced approaches, such as StyleGAN2, Morph-PIPE and MorCode}. In contrast, the generation of only frequency-space artifacts performs poorly, in terms of all three evaluation metrics. However, when combined with pixel-level artifacts, frequency-space artifacts significantly reduce the EER and BPCER@APCER performance in most of the morphing attack categories. This finding underscores the advantage of our method in leveraging both artifact generation strategies to enhance overall detection performance in diverse morphing attack scenarios.


\vspace{1mm}\noindent\textbf{Explainability Analysis.} To gain deeper insight into SelfMAD, we generate Grad-CAM~\cite{gradcam_iccv_2017} heatmaps, that show the regions of interest identified by the model during the classification of bona fide and morphing attack samples. For this purpose, we use the best-performing SelfMAD backbone, HRNet-W18, and extract heatmaps at the fourth stage of the classifier. Selected examples are illustrated in Fig.~\ref{fig:heatmaps}. We observe that SelfMAD primarily focuses on the eyes, nostrils and borders of the lips. These regions are in fact facial parts where pixel-space irregularities, such as ghosting artifacts, are most apparent. Similarly, StyleGAN2-generated face images are also know to exhibit inconsistencies in these areas, such as asymmetries, unnatural reflections, blurry edges or a lack of clear edge definition etc. Additionally, in some images the model also highlights hairy regions such as beard, mustache, eyebrows and hairstyles (Fig.~\ref{fig:heatmaps}, row 4).

\vspace{-5pt}
\section{Conclusion}
We presented SelfMAD, a self-supervised model that detects face morphing attacks by replicating typical artifacts of various common face morphing techniques. In extensive experiments across multiple widely used datasets, SelfMAD consistently outperformed state-of-the-art MAD models, reducing the equal error rate (EER) by a significant margin compared to leading discriminative and unsupervised MAD approaches. By leveraging both pixel and frequency-space artifacts generation, SelfMAD demonstrated superior generalization, particularly against unseen morphing techniques.

\vspace{1mm}\noindent\textbf{Limitations and Future Work.} Despite its strong performance, SelfMAD has some limitations. While the model shows robustness across a range of face morphing techniques, \hl{including some very recent GAN and diffusion-based attacks}, its detection performance slightly decreases when confronted with face morphs produced StyleGAN2. To address this issue, our future work will focus on enhancing the augmentation techniques used to simulate morphing artifacts, incorporating additional proxy tasks to further boost the model’s performance. Moreover, we plan to refine the feature extraction architecture and further expand our evaluation experiments by including additional recent face morphing attack methods to provide an even broader and more comprehensive assessment of our approach.




\vspace{-5pt}
\section{Ethical Impact Statement} 

The primary goal of this work is to enhance the robustness and generalization of Morphing Attack Detection (MAD) techniques in order to protect face recognition systems from identity fraud and other security threats. The presented research does not involve human subjects, and all datasets utilized are publicly available, ensuring compliance with standard ethical practices in data handling.

We acknowledge that the use of facial recognition technologies has raised concerns about privacy, surveillance, and potential misuse in societal contexts. However, the proposed SelfMAD method does not pose significant privacy concerns, as it is designed exclusively to detect manipulated face images (morphs) rather than perform face recognition. The development and deployment of MAD systems enhance the security of face recognition technologies, helping prevent identity theft and fraud by ensuring individuals cannot use morphed images to falsely authenticate themselves.

While MAD technology has clear benefits in securing digital identity systems, we acknowledge that there is a possibility for unintended misuse if deployed without adequate oversight. For example, in contexts where MAD tools are used, improper configurations or interpretations of results could lead to incorrect conclusions, such as misidentifying legitimate users as fraudulent due to false positives. To mitigate such risks, we recommend adherence to strict operational protocols, including continuous performance evaluation, calibration for different datasets, and clear guidelines on the application scope.

\hl{It is important to note that the standard public datasets used in this research, while widely accepted in the research community, may carry inherent biases, such as imbalanced demographics, which could influence the performance of the proposed model. In this work, we adhered to established protocols from prior studies to ensure consistency and comparability. However, we emphasize the need for future research to incorporate more diverse datasets and benchmarks that better reflect real-world scenarios, reducing potential biases and enhancing fairness in MAD systems.}






{\small
\bibliographystyle{ieee}
\bibliography{bibliography}
}

\end{document}